\documentclass[letterpaper, 10 pt, journal, twoside]{ieeetran}
\usepackage[english]{babel}
\usepackage{amsmath}
\usepackage{amssymb}
\usepackage{booktabs}
\usepackage{siunitx}
\usepackage{graphicx}
\usepackage{multirow}
\usepackage{amsfonts}
\usepackage{enumerate}
\usepackage{tabularx}
\usepackage{algorithm,algorithmic}
\usepackage{bm}
\usepackage{xcolor}
\usepackage{mdframed}
\usepackage{multirow}
\usepackage{newunicodechar}
\usepackage{adjustbox}
\usepackage{authblk}
\usepackage{color}
\usepackage{xurl}
\usepackage{subcaption}
\usepackage{cite}
\makeatletter
\let\NAT@parse\undefined
\makeatother
\usepackage{hyperref}
\usepackage{relsize}
\usepackage{float}
\usepackage{pifont}

\hypersetup{
    colorlinks=true,
    linkcolor=blue,
    filecolor=magenta,      
    urlcolor=cyan,
    pdfpagemode=FullScreen,
}

\begin{document}

\title{HoLoArm: Deformable Arms for Collision-Tolerant Quadrotor Flight}

\author{Quang Ngoc Pham$^{1}$, Jonas Eschmann$^{2}$, Yang Zhou$^{3}$, Alejandro Ojeda Olarte$^{3}$,\\ Giuseppe Loianno$^{2*}$, and Van Anh Ho$^{1*}$

\thanks{Manuscript received: August, 31, 2025; Revised: November, 28, 2025; Accepted: December, 30, 2025.}%Use only for final RAL version
\thanks{This paper was recommended for publication by
Editor Yong-Lae Park upon evaluation of the Associate Editor and Reviewers comments. This work was supported by the NSF CAREER Award 2546659, the NSF CPS Grant CNS-2121391, the DARPA YFA Grant D22AP00156-00, JSPS KAKENHI Grant Number 24KK0082, JST CREST Grant Number JPMJCR2554.}        
\thanks{$^{1}$The authors are with Japan Advanced Institute of Science and Technology (JAIST), Ishikawa, 923-1292 Japan. Email:
        {\tt\footnotesize \{pham\_n\_quang, van-ho\}@jaist.ac.jp.}}
\thanks{$^{2}$The authors are with the University of California Berkeley,
Department of Electrical Engineering and Computer Sciences,
Berkeley, CA 94720, USA. Email:{\tt\footnotesize \{jonas.eschmann,loiannog\}@eecs.berkeley.edu}.}
\thanks{$^{3}$The authors are with the New York University, Tandon School of Engineering, Brooklyn, NY 11201, USA. Email: {\tt\footnotesize \{yangzhou, ao2709\}@nyu.edu}.}%

\thanks{$^{*}$Corresponding authors. }
\thanks{Digital Object Identifier (DOI): 10.1109/LRA.2026.3656783}}

\markboth{IEEE Robotics and Automation Letters. Preprint Version. Accepted December, 2025}
{Pham \MakeLowercase{\textit{et al.}}: HoLoArm: Deformable Arms for Collision-Tolerant Quadrotor Flight} 

\maketitle
\IEEEpubid{\begin{minipage}[t]{\textwidth}\ \\[10pt]
{\copyright 2026 IEEE. Personal use is permitted, but republication/redistribution requires IEEE permission.
See \url{https://www.ieee.org/publications/rights/index.html} for more information.}
\end{minipage}}

\begin{abstract}
The increasing use of drones in human-centric applications highlights the need for designs that can survive collisions and recover rapidly, minimizing risks to both humans and the environment. We present \textit{HoLoArm}, a quadrotor with compliant arms inspired by the \textit{nodus} structure of dragonfly wings. This design provides natural flexibility and resilience while preserving flight stability, which is further reinforced by the integration of a Reinforcement Learning (RL) control policy that enhances both recovery and hovering performance. Experimental results demonstrate that \textit{HoLoArm} can passively deform in any direction, including axial one, and recover within 0.3-0.6\,s depending on the direction and level of the impact. The drone can survive collisions at speeds up to 7.6\,m/s and carry a 540\,g payload while maintaining stable flight. This work contributes to the morphological design of soft aerial robots with high agility and reliable safety, enabling operation in cluttered and human shared environments, and lays the groundwork for future fully soft drones that integrate compliant structures with intelligent control.
\end{abstract}


%%%%%%%%%%%%%%%%%%%%%%%%%%%%%%%%%%%%%
%%%%%%%%%%%%%%%%%%%%%%%%%%%%%%%%%%%%%%%
\section{Introduction} 
%%%%%%%%%%%%%%%%%%%%%%%%%%%%%%%%%%%%%%%
%background
\IEEEPARstart{M}{echanical} safety of Unmanned Aerial Vehicles (UAVs) represents a critical challenge when operating in dynamic environments where physical interactions with obstacles are frequent. While many solutions have focused on collision avoidance through sensing and path planning, collisions often remain a fundamental aspect of real-world operations for several reasons. They are inevitable due to limitations in sensing and planning. They can be intentionally exploited in some applications, such as tactile-based navigation~\cite{Bredenbeck2025} or human-assisted physical interaction for assisted drone piloting~\cite{Sondoqah2024}. They are required for physical interaction with the environment, such as landing or contact-based tasks. In all these cases, UAVs require collision-tolerant designs to ensure structural integrity and mission success. For instance, vision-based approaches may fail under dense fog or heavy rain~\cite{Bessonov2025, RandieriC2025}; LiDAR or ultrasound-based sensing can become unreliable in environments with strong acoustic absorption~\cite{Dreissig2023, H.andKartsch2024}, and path planning strategies alone may be insufficient under high wind conditions~\cite{Oettershagen2017}. Despite these limitations, relatively few studies have emphasized proactive mechanical design to mitigate the effects of collisions when they occur. This motivates our recognition that research on structural softening of drone frames to minimize damage to the lowest possible level is both necessary and timely.

\begin{figure}[t]
\centering
\includegraphics[width=0.9\columnwidth]{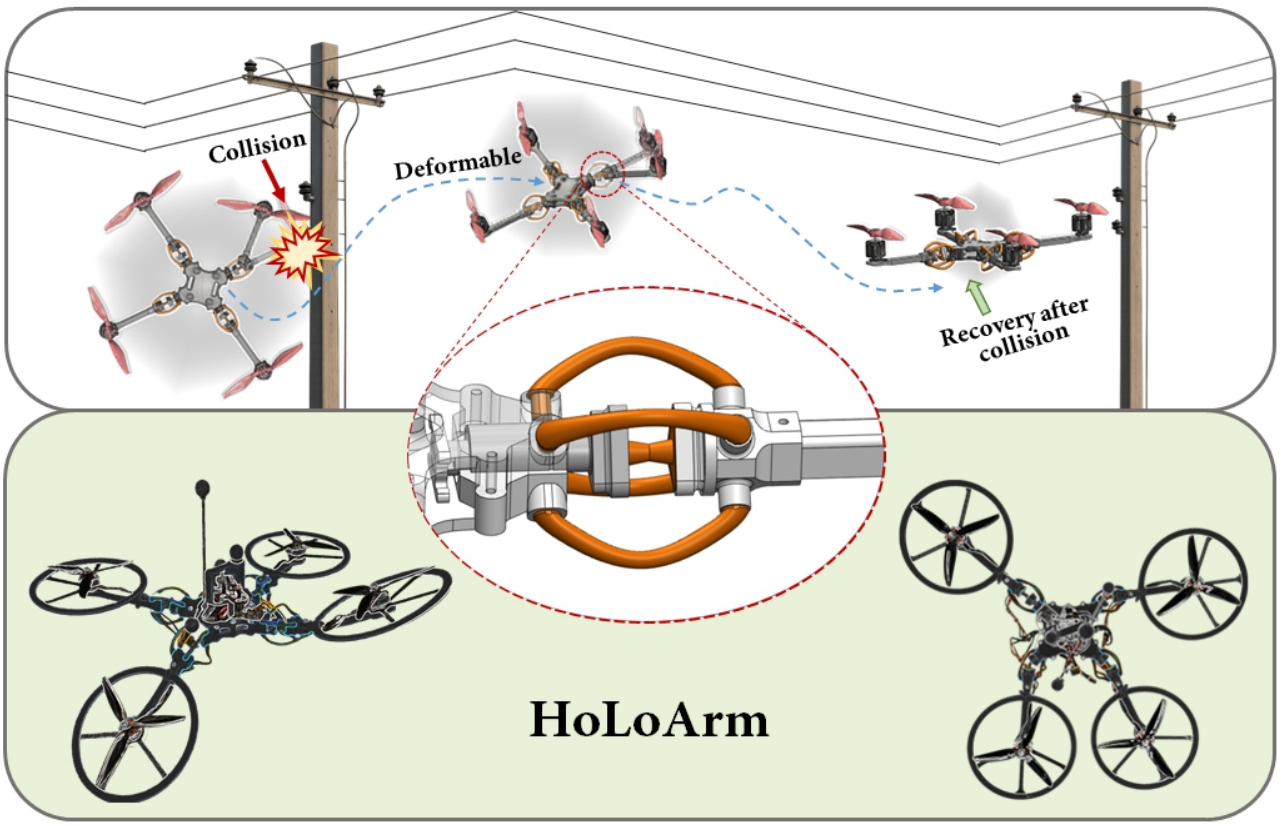}
\caption{Illustration of the visionary use of the \textit{HoLoArm} with deformable joint arms, showcasing its ability to recover after a collision.}
\label{fig:HoLo-arm}
\end{figure}

Several recent designs have shown promising potential. The SoBAR platform, a fully soft-bodied aerial robot, employs pneumatic actuation to modulate stiffness and demonstrates recovery capability under multidirectional collisions ~\cite{nguyen2022sobar}. Another study analyzed the placement of compliant elements within the airframe, revealing that softening the internal frame rather than only using external protective layers enhances impact energy absorption and prolongs contact duration, thus reducing peak collision accelerations ~\cite{abazari2024compliance}. In addition, research on pneumatic morphing for soft quadrotors has demonstrated the ability to alter flight geometry through pneumatic actuation; however, a thorough assessment of the nonlinear interactions between soft arm forces, propeller downwash, and overall flight stability remains necessary~\cite{sumathy2025pneumatic}.
%\\\\\\\\\\\\\\\\\\\\\\\\\\\\\\\\\\\\\\\\\\\\\\\\\

%\\\\\\\\\\\\\\\\\\\\\\\\\\\\\\\\\\\\\\\\\\\\\\\\\\\\\\

Nature-inspired designs, bio-morphic mechanisms have been developed to allow drones to actively deform their shape to pass through narrow gaps or avoid obstacles~\cite{8794373,10.1007/978-3-319-95972-6_42}. Other designs focus on adaptive scaling, where the drone can morph between maximum and minimum physical shape during flight~\cite{MorphoCopter2025, falanga2018ral}. Several efforts have adopted soft materials to passively absorb collision energy and improve resilience. For instance, one research introduced a compliant propeller to withstand impact~\cite{Tombo}, while \emph{Morphy} drone in~\cite{DePetris2024Morphy} features soft quadcopter arms to minimize structural damage. 

%In this paper, we propose a soft-armed quadrotor platform inspired by the biomechanics of the hinge-like joint found in flying insects, called \textit{HoLoArm} (Fig.~\ref{fig:HoLo-arm}), which is able to accommodate propeller sizes ranging from 6 to 13\,inches. \textcolor{blue}{The soft joint in our design is functionally inspired by the nodus region found in insect wings, which serves as a compliant hinge that enables localized deformation and passive impact absorption~\cite{Wootton1992}. Our implementation adopts these functional principles rather than replicating the biological geometry.} By integrating mechanical compliance into the arm structure, our drone can absorb impact forces more effectively while maintaining stable post-collision flight. The modular arm design also facilitates the ease of assembly and repair. We highlight the fabrication methodology and experimentally compare the collision behavior of soft versus rigid drones under identical test conditions to demonstrate enhanced survivability and recovery. For a novel design that combines softness with structural complexity, deriving an accurate mathematical model for the entire drone frame remains a major challenge. This difficulty arises from inevitable approximation errors in capturing the nonlinear deformation of soft components, as well as the increased complexity associated with modeling coupled interactions between rigid and compliant elements.
In this paper, we propose a soft-armed quadrotor platform inspired by the biomechanics of the hinge-like joint found in flying insects, called \textit{HoLoArm}  (Fig.~\ref{fig:HoLo-arm}). The soft joint in our design is functionally inspired by the nodus region found in insect wings, which serves as a compliant hinge that enables localized deformation and passive impact absorption~\cite{Wootton1992}. Our implementation adopts these functional principles rather than replicating the biological geometry. \textit{HoLoArm} can absorb multi-directional impacts while accommodating a wide range of propeller sizes (6–13 inches). The modular arm structure also simplifies fabrication and repair, opening a promising direction for developing deformable drones with enhanced collision survivability. However, accurately modeling the full dynamics of such a hybrid soft–rigid structure remains highly challenging due to nonlinear deformation and complex interactions between compliant and rigid components. These difficulties make traditional model-based control approaches less effective. To overcome this limitation, we employ a reinforcement learning (RL) policy to control \textit{HoLoArm}, leveraging its ability to operate without requiring an accurate mathematical model and to adapt naturally to hardware complexity and structural compliance. We present the fabrication process, RL training implementation, and experimental comparisons between soft and rigid drones under identical impact tests, demonstrating that \textit{HoLoArm} achieves better resilience and post-collision recovery.

The structure of this paper is organized as follows. 
Section~\ref{sec:safety_control_strategy} reviews the related work. 
Section~\ref{sec:Hard and Soft Design} details the hardware design and the control software architecture. 
Section~\ref{sec:Experimental} presents the experimental results, 
and Section~\ref{Conclution} concludes the paper.

%%%%%%%%%%%%%%%%%%%%%%%%%%%%%%%%%%%%%%%
\section{Related Works} \label{sec:safety_control_strategy}
%%%%%%%%%%%%%%%%%%%%%%%%%%%%%%%%%%%%%%%
In recent years, mechanical solutions to enhance UAV collision tolerance have received significant attention. One common approach is the design of morphing wings or foldable arms to reduce collision risks in confined environments \cite{zhao2017deformable}. However, most of these solutions require additional actuators such as servos \cite{desbiez2017xmorf}, \cite{riviere2018agile} or passive mechanical joints \cite{nguyen2022compliance}, which increase system weight and complexity, while still relying on rigid components that limit safe interaction with the environment. Another line of research focuses on protective structures. Protective cages \cite{klaptocz2013euler} are a typical example, valued for their simplicity and low cost. However, they are often bulky and heavy, reducing flight efficiency and even introducing new risks. An alternative strategy involves the use of soft materials in the airframe to absorb impact energy. While fully soft-bodied UAVs have been studied, they face challenges in control and energy efficiency. More practical approaches selectively soften critical parts of the airframe, such as upper arms or joints. For instance, \cite{Liu2021ImpactResilientQuadrotor} proposed a spring-based mechanism to absorb axial impacts, though it remains limited when subjected to perpendicular forces. The Morphy platform introduced soft joints made from Elastic Resin 50A \cite{DePetris2024Morphy}, enabling UAV arms to flex and recover after multidirectional collisions. However, this approach is mostly suited to small-scale UAVs, and its ability to absorb axial forces remains limited, especially for larger drones with bigger propellers. \emph{Morphy} represents an important advancement in compliant aerial robots, demonstrating soft joints capable of passive deformation and active shape adaptation. Its morphology-aware sensing and control enable navigation through confined spaces and resilience to moderate impacts. Although soft-material designs are promising, balancing collision tolerance with system efficiency remains challenging. This motivates our quadrotor design, which employs high-elasticity TPU arms to absorb multidirectional impacts including axial forces while maintaining structural simplicity and minimal added weight.

In this work, the \textit{HoLoArm} platform targets a different objective from \emph{Morphy} by maximizing passive mechanical impact resilience. The nodus-inspired compliant joint enables effective impact absorption and rapid recovery without embedded sensors or morphing mechanisms, resulting in a simpler, more durable, and easily repairable structure that scales to larger propeller sizes. By avoiding sensors or heterogeneous materials inside the joints, our design preserves structural uniformity and reduces complexity. Furthermore, \textit{HoLoArm} is engineered to withstand multidirectional impacts, particularly axial loads in a direction where Morphy’s monolithic soft structure remains limited. 

The main contributions of this work are summarized as follows:
\begin{itemize}
    \item \textbf{Bio-inspired design:} A soft-armed quadrotor inspired by insect hinge-like joints, integrating mechanical compliance to enhance impact resilience.
    \item \textbf{Scalable and modular structure:} A modular arm design adaptable to propellers from 6 to 13\,inches, simplifying assembly and maintenance.
     \item \textbf{Control integration:} Application of a Reinforcement Learning (RL) policy for stable autonomous flight.
    \item \textbf{Experimental validation:} Comparative tests between soft and rigid drones under identical collision conditions, demonstrating superior survivability and recovery.
    
\end{itemize}
\section{Hardware and Software Design} \label{sec:Hard and Soft Design}
\subsection{Nodus-Inspired Design in \textit{HoLoArm }} 
%\\\\\\\\\\\\\\\\\\\\\\\\\\\\\\\\\\\\

\begin{figure}[t]
\centering
\includegraphics[width=0.9\columnwidth]{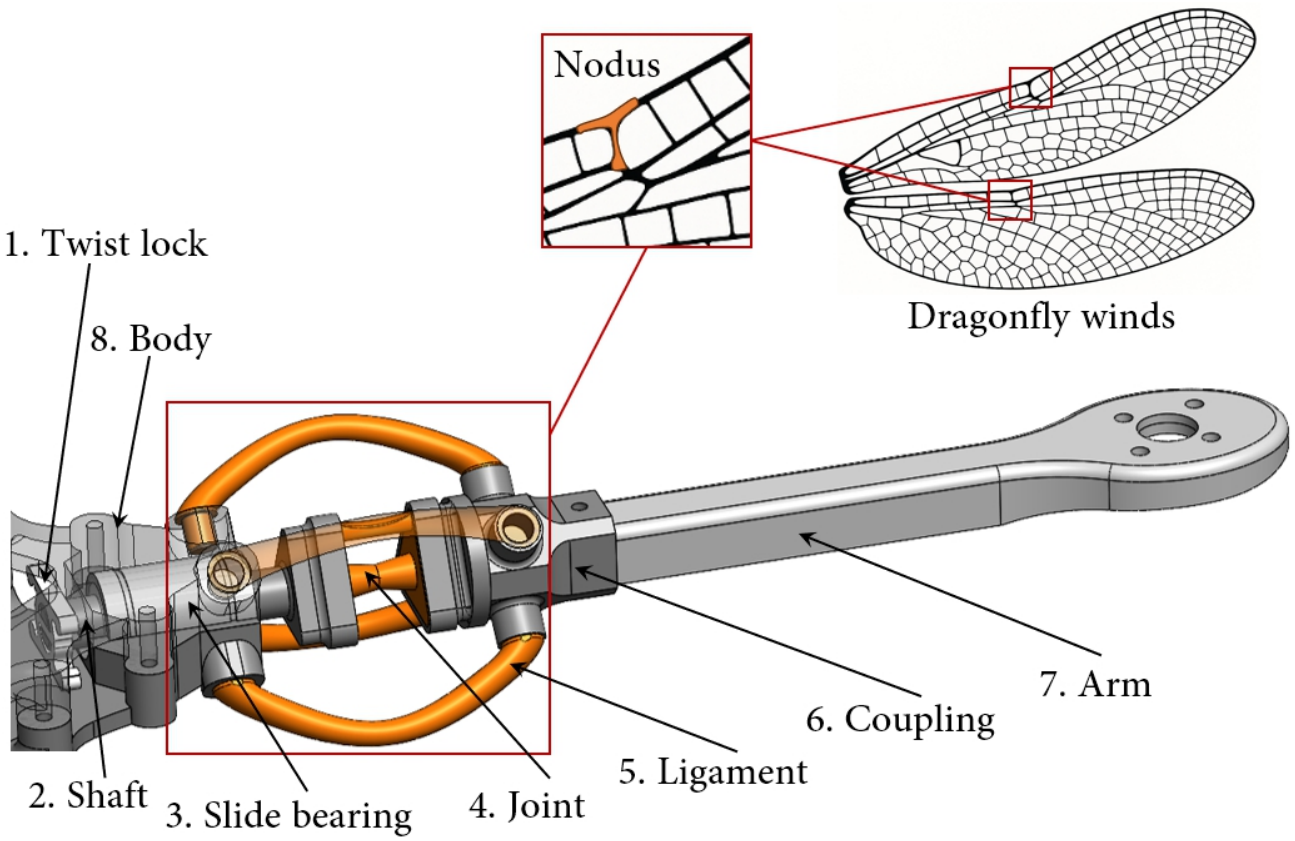}
\caption{Bioinspired design concept based on the dragonfly’s nodus structure.}
\label{fig:fig_nodus}
\end{figure}

%\vspace{-2mm}
%\\\\\\\\\\\\\\\\\\\\\\\\\\\\\\\\\\\\
%\\\\\\\\\\\\\\\\\\\\\\\\\\\\\\\\\\\\
%\\\\\\\\\\\\\\\\\\\\\\\\\\\\\\\\\\\\
In nature, winged insects have evolved body structures that allow them to resist shocks or absorb impacts, thus minimizing damage during collisions~\cite{Mountcastle2014}. In particular, when studying dragonflies in depth, biologists discovered a crucial component in their wing structure called the {nodus}, a hinge-like structure that can passively bend upon impact or aerodynamic disturbances without causing structural failure. With a combination of flexible \textit{resilin} material and stiff vein regions, the nodus creates a {hybrid structure that is flexible and robust}, enabling dragonflies to operate efficiently while maintaining resilience to environmental forces~\cite{Wootton1992,Rajabi2020}.

Inspired by this unique biological feature, we propose the design of a {soft arm with a passively deformable joint}, modeled after the operating principle of the nodus. The flexible joint section of the arm is designed to absorb impact forces and deform in all directions, improving the system’s resilience and safety in the event of unexpected collisions during flight. Fig.~\ref{fig:fig_nodus} illustrates the detailed nodus-inspired structural design integrated into the \textit{HoLoArm} drone. The structure consists of {eight interlinked components}, securely assembled using Loctite 416 Instant Adhesive (cyanoacrylate glue) and {M3 x 18\,mm screws}. Each element has a specific function and, when combined, forms a complete mechanism that enables \textit HoLoArm to adapt to external forces and recover safely to continue its flight tasks. Here, each nodus incorporates a sliding bearing LM88U , which allows the shaft (shown green in Fig. ~\ref{fig:fig_nodus}) to slide horizontally along the body. This sliding capability enables the drone to withstand impact forces along the longitudinal direction of the arm. Combined with the elasticity of the four ligaments, these forces are absorbed and gradually restored to the original position, thanks to the TPU material that makes up the ligaments. Beyond their role in passive recovery, the four ligaments also serve to secure the initial configuration of the structure, ensuring the \textit{HoLoArm } functions effectively. The joint is also made of TPU. It is designed asymmetrically to prevent the arm from sagging due to gravity when idle and to reduce upward bending caused by thrust forces during flight. At the center of the main axis of the blue part, we specifically designed a narrower diameter segment to allow the mechanism to bend flexibly like a ball joint when subjected to off-axis impact forces. The coupling not only ensures symmetrical placement of the four ligaments for uniform restoring forces, but is also designed for easy assembly and disassembly of the motor-carrying arm, making the \textit{HoLoArm } more convenient for transport. The four orange components are curved to provide built-in elasticity, and their placement enables the nodus structure to respond and recover from multi-directional impacts on the \textit{HoLoArm }. Lastly, the twist lock solves the issue of thrust force-induced rotation around the arm’s longitudinal axis during motor operation. This feature not only enhances flight stability, but also ensures that ligaments deformation occurs in the intended directions, as designed by the authors.
%\FloatBarrier

\begin{figure}[t]
\centering
\includegraphics[width=0.8\columnwidth]{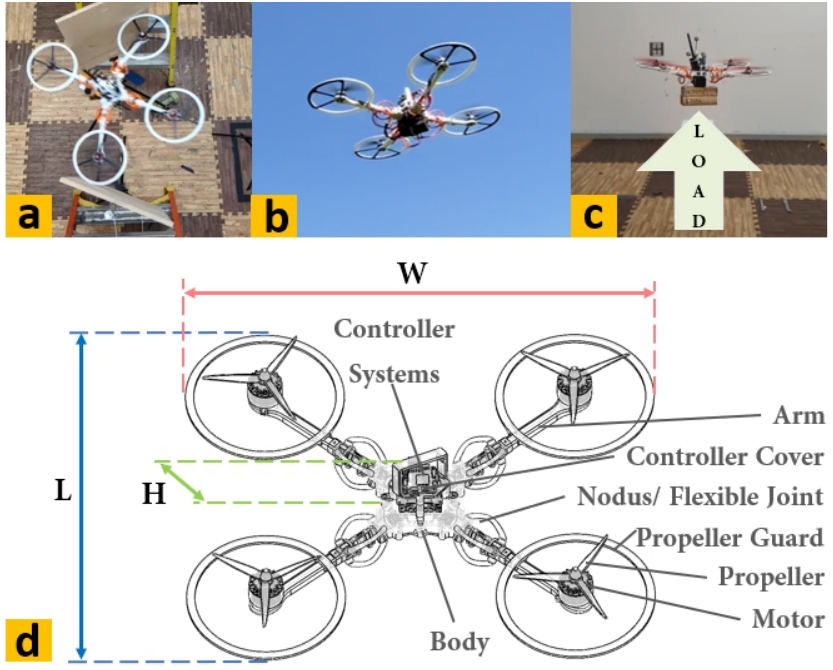}
\caption{Experimental demonstrations of the \textit{HoLoArm} in various scenarios: (a) Narrow-gap traversal test; (b) outdoor flight; (c) Lifting heavy payloads, and (d) Hardware configuration of the deformable drone.}
\label{fig:Fig-drone-dimension}
\end{figure}
\vspace{-2mm}
%The coupling not only ensures symmetrical placement of the four ligaments for uniform restoring forces, but is also designed for easy assembly and disassembly of the motor-carrying arm, making the \textit{HoLoArm } more convenient for transport. The four orange components are curved to provide built-in elasticity, and their placement enables the nodus structure to respond and recover from multi-directional impacts on the \textit{HoLoArm }.

%Lastly, the twist lock solves the issue of thrust force-induced rotation around the arm’s longitudinal axis during motor operation. This feature not only enhances flight stability, but also ensures that ligaments deformation occurs in the intended directions, as designed by the authors.
%\FloatBarrier
%%%%%%%%%%%%%%%%%%%%%%%%%%%%%%%%%%%%%%%%%%%
%%%%%%%%%%%%%%%%%%%%%%%%%%%%%%%%%%%%%%%%%%%
\subsection{HoLoArm's airframe}

%\\\\\\\\\\\\\\\\\\\\\\\\\\\\\\\\\\\\
%\\\\\\\\\\\\\\\\\\\\\\\\\\\\\\\\\\\\
%\\\\\\\\\\\\\\\\\\\\\\\\\\\\\\\\\\\\

The design of \textit{HoLoArm} is not only aimed at tackling collision challenges in lab environments (Fig.~\ref{fig:Fig-drone-dimension}a), but also at enabling real-world applications such as outdoor operation (Fig.~\ref{fig:Fig-drone-dimension}b) and heavy-payload lifting (Fig.~\ref{fig:Fig-drone-dimension}c), making it well-suited for practical UAV tasks. To support this design goal, we focus on developing a drone with a robust structure that meets the demands of modern UAV applications. Figure~\ref{fig:Fig-drone-dimension}d illustrates the drone’s dimensions, components, and structural layout. Even the structure can be scaled, in this paper, we aim for the 9-inch propeller adopted drone. The overall size of the proposed \textit{HoLoArm} drone is $458~\times~512~\times~52$\,mm (L~$\times$~W~$\times$~H) when the arms are in their nominal configuration, with a total weight of 970\,g of the drone including the battery.

To ensure a balance between flexibility and structural rigidity, various materials are used in different components. The ligaments and joints, which require elasticity, are 3D-printed using TPU. In contrast, the arms, main body, and controller cover are fabricated using PLA, providing the stiffness and mechanical strength necessary for flight stability. The propeller guards are made from a composite of carbon fiber and PLA, offering a compromise between mechanical robustness, lightweight design, and cost-effectiveness.

For propulsion, \textit{HoLoArm} drone is equipped with four T-MOTOR Ultralight P2505 1850\,KV brushless motors paired with 6032 propellers, and powered by a 4-cell 5000\,mAh LiPo battery. This configuration delivers the required thrust and overall performance for the targeted tasks. The drone is controlled using a Pixracer Pro flight controller running the PX4 firmware, integrated with a Speedybee 60A BLHeli\_S 4\_in\_1 ESC to ensure reliable flight control and power management.

\begin{figure}[t]
\centering
\includegraphics[width=0.8\columnwidth]{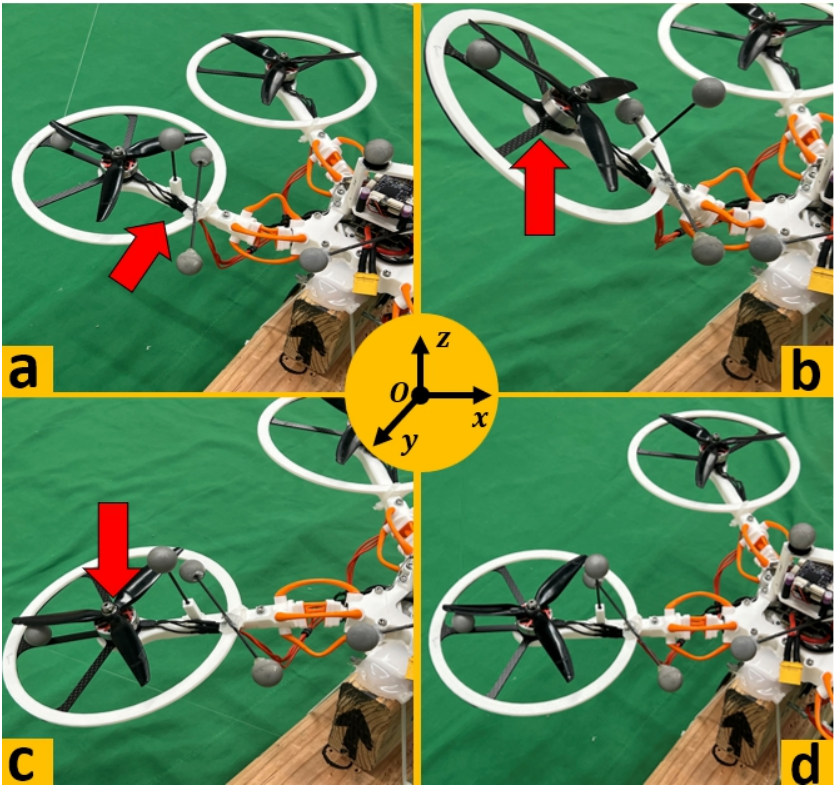}
\caption{\textit{HoLoArm} drone in various scenarios: (a) Lateral bending; (b) Upward bending; (c) Downward bending, and (d) Normal state.}
\label{fig:fig-drone-deformable}
\end{figure}
%\vspace{-2mm}

%%%%%%%%%%%%%%%%%%%%%%%%%%%%%%%%%%%%%%%%%%%
%%%%%%%%%%%%%%%%%%%%%%%%%%%%%%%%%%%%%%%%%%%
\subsection{Arm Response under Various Loading Scenarios:} 

%\\\\\\\\\\\\\\\\\\\\\\\\\\\\\\\\\\\\
%\\\\\\\\\\\\\\\\\\\\\\\\\\\\\\\\\\\\
%\\\\\\\\\\\\\\\\\\\\\\\\\\\\\\\\\\\\
Figure~\ref{fig:fig-drone-deformable} illustrates the pre- and post-deformation states of the \emph{joint} and \emph{ligament} structures under various loading scenarios, highlighting their mechanical response. Lateral forces acting perpendicular to the $zOx$ or $zOy$ planes (Fig.~\ref{fig:fig-drone-deformable}a).Upward forces perpendicular to the $xOy$ plane (Fig.~\ref{fig:fig-drone-deformable}b). Downward forces are also perpendicular to the $xOy$ plane (Fig.~\ref{fig:fig-drone-deformable}c). Additionally,  the HoLoArm in its normal state is shown (Fig.~\ref{fig:fig-drone-deformable}d). Under these conditions, the arm bends to a deformation threshold determined by the magnitude of the applied force, then passively returns to its nominal shape due to the elastic properties of the TPU-based components. To characterize the deformation behavior of the soft arm and quantify the recovery time required to return to a safe state that enables continuous flight after external disturbances, we employ a Vicon\footnote{\url{www.vicon.com}} motion capture system to measure the relative motion of one arm with respect to the \emph{HoLoArm} drone body at a sampling frequency of 200\,Hz. The drone was fixed in place, and the arm was pulled in a desired direction using a thread, which was then released to allow the arm to recover freely toward its initial state. Each test was repeated three times, considering the same deformation condition, and the average result was used to minimize measurement errors. Referring to the recovery performance of the \emph{Morphy} drone (figure 8 in Morphy paper) ~\cite{DePetris2024Morphy}, we note that the published figure shows that the arm continues to exhibit a residual deviation for several seconds after the initial rapid recovery phase. While \emph{Morphy} demonstrates stable post-impact flight, the relatively slow convergence toward its nominal position suggests that prolonged deviation may influence the transient stability of the platform after a collision. Motivated by this observation, we define the recovery threshold for \emph{HoLoArm} as the moment when the arm angle returns to within 1\,deg of its original orientation. This criterion reflects a sufficiently small deviation for maintaining stable flight and allows us to quantitatively assess the responsiveness of the proposed soft-joint mechanism.
%\\\\\\\\\\\\\\\\\\\\\\\\\\\\\\\\\\\\\
%\\\\\\\\\\\\\\\\\\\\\\\\\\\\\\\\\\\\\
%\\\\\\\\\\\\\\\\\\\\\\\\\\\\\\\\\\\\\

\begin{figure}[t]
\centering
\includegraphics[width=1\columnwidth]{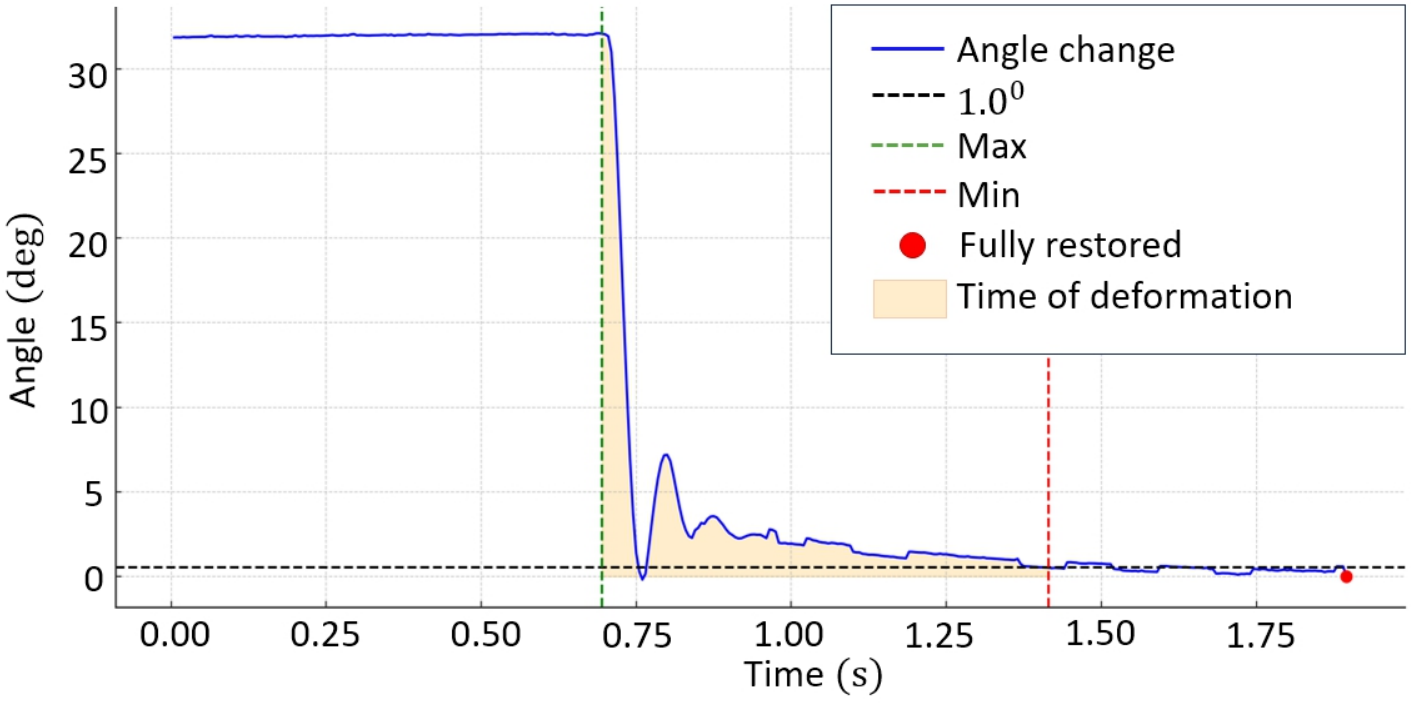}
\caption{Recovery vs. time of the soft arm when subjected to external forces acting perpendicular to the $xOz$ or $yOz$ planes.}
\label{fig:Graphic-XOY}
\end{figure}
%\vspace{-2mm}
%\\\\\\\\\\\\\\\\\\\\\\\\\\\\\\\\\\\\
%\\\\\\\\\\\\\\\\\\\\\\\\\\\\\\\\\\\\
%\\\\\\\\\\\\\\\\\\\\\\\\\\\\\\\\\\\\
As shown in Figure~\ref{fig:Graphic-XOY}, the green dashed line intersects the blue recovery profile at the point where the arm reaches its maximum deformation angle of approximately 32$^\circ$. The black dashed horizontal line intersects the vertical axis at 1$^\circ$, which is the threshold defined by the authors as the safe recovery state at which the \emph{HoLoArm} is deemed stable enough to resume its flight task. The red dashed vertical line marks the moment when the arm returns to within 1$^\circ$ of its original position, indicating successful recovery. The time elapsed between the peak deformation (green line) and the recovery threshold (red line) is approximately $0.72$\,seconds, under external forces applied perpendicular either to the $zOy$ or $zOx$ planes.
%\\\\\\\\\\\\\\\\\\\\\\\\\\\\\\\\\\\\\
%\\\\\\\\\\\\\\\\\\\\\\\\\\\\\\\\\\\\\
%\\\\\\\\\\\\\\\\\\\\\\\\\\\\\\\\\\\\\

\begin{figure}[t]
\centering
\includegraphics[width=0.9\columnwidth]{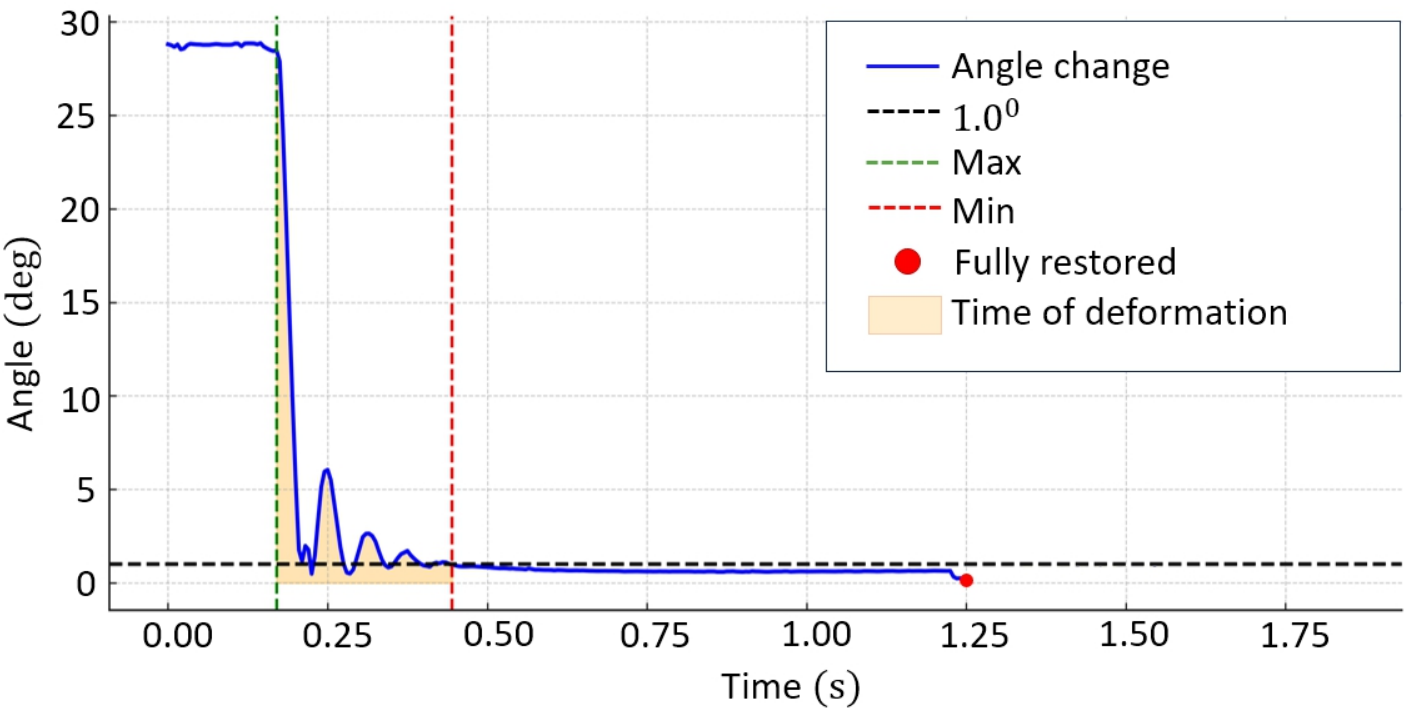}
\caption{Recovery over time of the soft arm when subjected to external forces acting upward to the  $xOy$ plane.}
\label{fig: Graphic-XOZ+}
\end{figure}
%\vspace{-2mm}
%\\\\\\\\\\\\\\\\\\\\\\\\\\\\\\\\\\\\
%\\\\\\\\\\\\\\\\\\\\\\\\\\\\\\\\\\\\
%\\\\\\\\\\\\\\\\\\\\\\\\\\\\\\\\\\\\
As shown in Figure~\ref{fig: Graphic-XOZ+}, we calculate the recovery time of the arm when subjected to an external force applied perpendicular to the $xOy$ plane along the positive direction of the $z$-axis. The measurement begins at $0.17$\,s, and the arm reaches the defined recovery threshold of $1$\,deg at $0.45$\,s. Thus, it takes approximately $0.27$\,s for the \textit{HoLoArm } to recover to the required position after having deviated by $28$\,deg from its original configuration.

\begin{figure}[b]
\centering
\includegraphics[width=0.9\columnwidth]{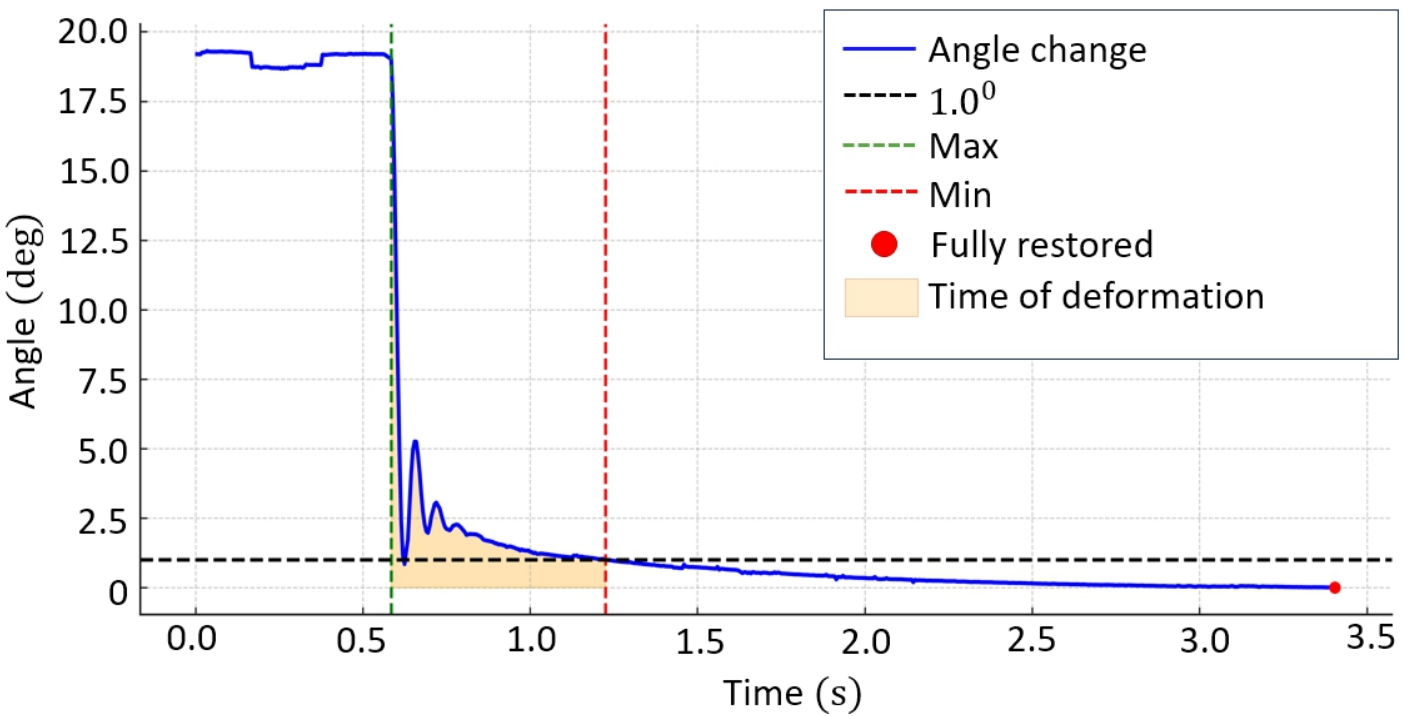}
\caption{Recovery over time of the soft arm when subjected to external forces acting downward to the  $xOy$ plane.}
\label{fig: Graphic-XOZ+2}
\end{figure}

As shown in Fig.~\ref{fig: Graphic-XOZ+2}, the arm experiences a $19$\,deg displacement after being subjected to an impact force applied perpendicular to the $xOy$ plane along the negative direction of the $Z$-axis. In this case, the \emph{HoLoArm} requires approximately $0.62$\,s for the arm to return to its predefined safe configuration.
%\\\\\\\\\\\\\\\\\\\\\\\\\\\\\\\\\\\\\
%\\\\\\\\\\\\\\\\\\\\\\\\\\\\\\\\\\\\\
%\\\\\\\\\\\\\\\\\\\\\\\\\\\\\\\\\\\\\

%\vspace{-2mm}
%\\\\\\\\\\\\\\\\\\\\\\\\\\\\\\\\\\\\
%\\\\\\\\\\\\\\\\\\\\\\\\\\\\\\\\\\\\
%\\\\\\\\\\\\\\\\\\\\\\\\\\\\\\\\\\\\

The red dots appearing in the plots of Figs.~\ref{fig:Graphic-XOY} and~\ref{fig: Graphic-XOZ+} and \ref{fig: Graphic-XOZ+2} indicate the positions corresponding to 0\,deg deviation. This clearly demonstrates that the soft structure of the arms in \emph{HoLoArm} is fully capable of returning to its original position after experiencing impact forces from various directions. Both the maximum deformation angle and the recovery time differ significantly across cases, highlighting the system’s ability to exhibit direction-dependent responses. These variations are meaningful, as they reflect the influence of structural differences, positional configurations, and, in particular, the asymmetric design of the \emph{Joint} component.
%\\\\\\\\\\\\\\\\\\\\\\\\\\\\\\\\\\\\\
%\\\\\\\\\\\\\\\\\\\\\\\\\\\\\\\\\\\\\
%\\\\\\\\\\\\\\\\\\\\\\\\\\\\\\\\\\\\\

\begin{figure}[t]
\centering
\includegraphics[width=0.8\columnwidth]{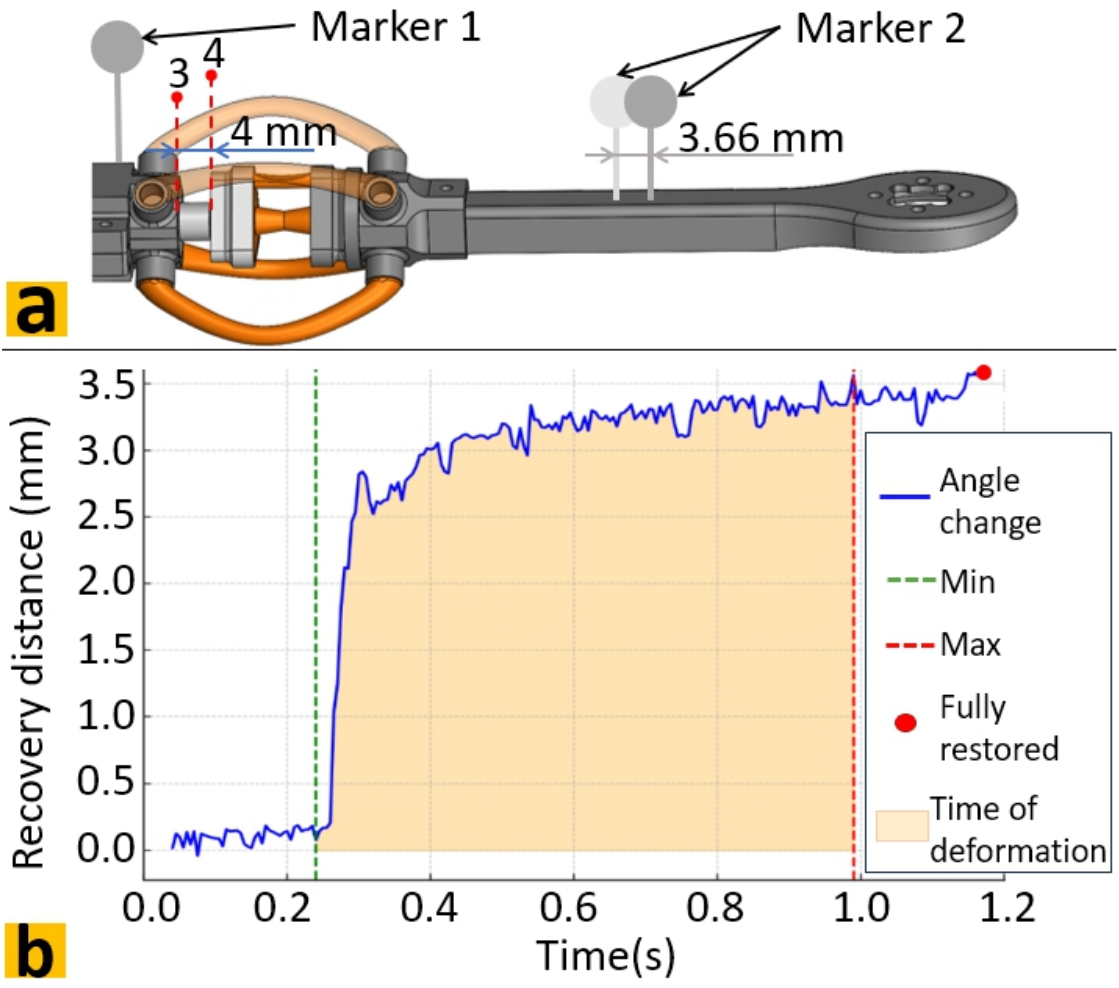}
\caption{Over-time recovery behavior of the soft arm subjected to an axial compression: (a) Design illustration of the experimental setup used to evaluate the arm’s recovery behavior along its longitudinal axis; (b) Measured recovery displacement of the arm along the longitudinal axis.}
\label{fig: Graphic-XYZ}
\end{figure}
%\vspace{-2mm}
%\\\\\\\\\\\\\\\\\\\\\\\\\\\\\\\\\\\\
%\\\\\\\\\\\\\\\\\\\\\\\\\\\\\\\\\\\\
%\\\\\\\\\\\\\\\\\\\\\\\\\\\\\\\\\\\\
When subjected to arm axial loads, the arm's recovery behavior is dictated solely by the intrinsic elastic configuration of the four ligaments without the joint part, which are symmetrically distributed around its structure.
When an axial impact force is applied along the arm’s longitudinal axis, the arm slides inward toward the main body through the embedded linear sliding mechanism. After absorbing the impact, the symmetrically arranged elastic ligaments surrounding the arm generate restoring forces that gradually return the arm to its original position. Fig.~\ref{fig: Graphic-XYZ}(a) presents the experimental procedure used to characterize the arm’s recovery behavior along its longitudinal axis. In this test, the arm was pulled rearward until point~4 reached geometric alignment with point~3, after which it was held momentarily before being abruptly released, allowing the arm to return to its initial configuration through its intrinsic elastic response. Marker~1 was rigidly attached to the \emph{HoLoArm} body to serve as a fixed reference, enabling the motion-tracking system to accurately quantify the displacement of Marker~2 mounted on the arm. While the theoretical design displacement is \SI{4.00}{\milli\meter}, the experimentally measured displacement of Marker\,2 was \SI{3.66}{\milli\meter}. As shown in Fig.~\ref{fig: Graphic-XYZ}(b), the measured recovery displacement is \SI{3.66}{\milli\meter}, occurring over a period of \SI{0.75}{\second}. The green curve intersects the blue characteristic curve at \SI{0.24}{\second}, 
corresponding to the instant at which the distance between the arm and the body reaches its minimum. The recovery point is indicated by the red guideline at \SI{0.97}{\second}, intersecting the characteristic curve at a distance of \SI{3.66}{\milli\meter}. Toward the final stage of the arm’s recovery process, continuous variations in the arm’s position within a narrow range can be observed, highlighting the presence of residual micro-oscillations.. These oscillations provide a clear representation of the system’s dynamic behavior: the restoring forces from the ligaments diminish over time, while the slight mechanical backlash introduced by the linear bearing contributes to the observed fluctuations. This combination of compliant ligament response and guided linear sliding enables the arm to dissipate impact energy effectively while ensuring stable re-centering performance.

Overall, these results collectively demonstrate that the \emph{HoLoArm} is capable of reliably recovering from large angular deformations (up to 32\,deg) and axial compression, within sub-second timescales, depending on the direction of the applied force. This rapid recovery capability is crucial for maintaining stable flight performance following collisions or disturbances.

\subsection{Reinforcement Learning Control Policy}
The soft structure of the proposed quadrotor introduces additional challenges in accurately capturing its dynamics compared to a rigid frame. Instead of relying on explicit system identification, we adopt a reinforcement learning (RL) policy that can adapt to uncertain and nonlinear dynamics. In the simulator, the model identifies and updates the dynamic parameters online, following the approach in~\cite{data-driven2024eschmann}. This enables stable control performance despite the complexity of the soft-body interactions.. The identified parameters are documented in Table \ref{table:dynamics-parameters}.

\begin{figure}[htbp]
\centering
\includegraphics[width=0.9\linewidth]{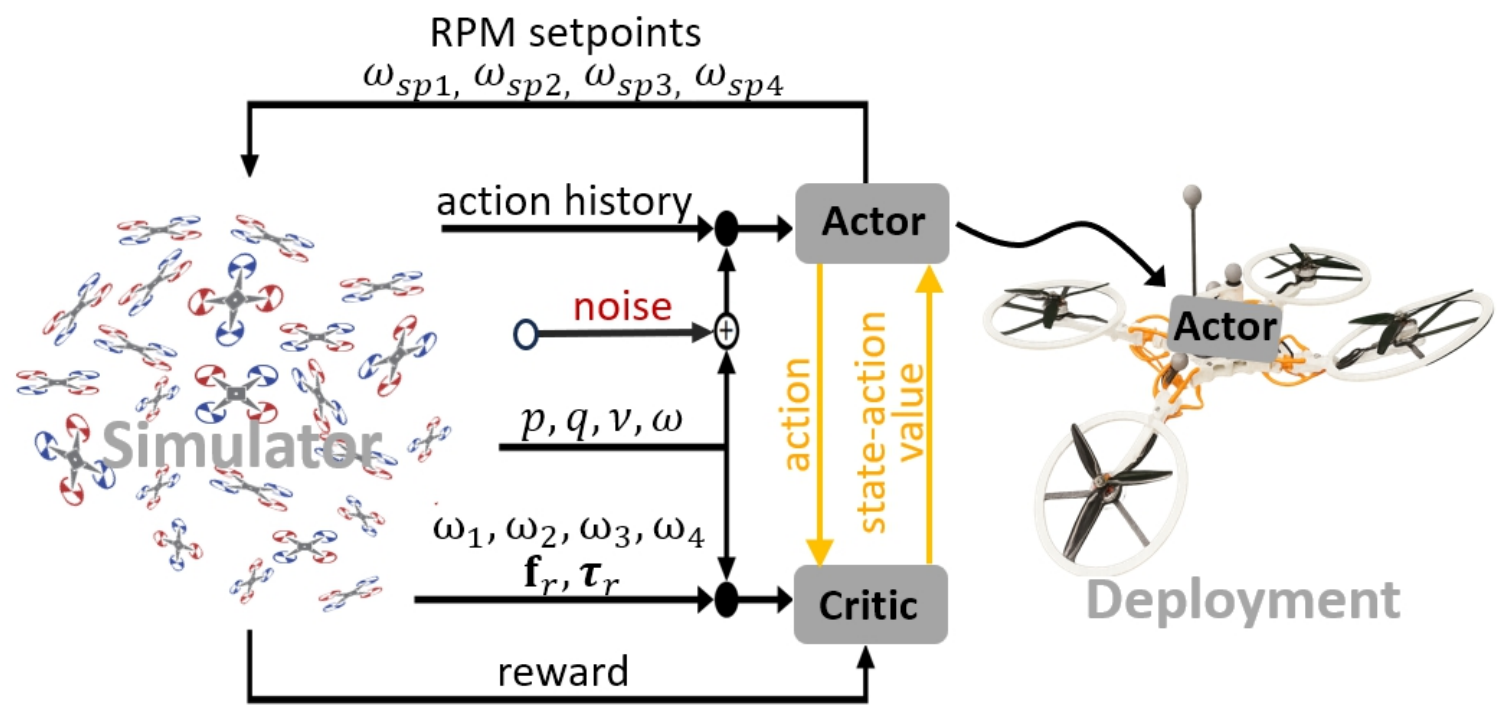}
\caption{The deep reinforcement learning setup. Here, $p$ is 3D position (global frame); $q$ is orientation quaternion; $v$ is linear velocity (global frame); $\omega$ is angular velocity (body frame).} 
\label{fig:overview}
\end{figure}
%\vspace{-2mm}

\begin{table}[t]
\centering
\caption{Identified dynamics parameters}
\begin{tabular}{| c | c |}
\hline
Parameter & Value \\
\hline
Mass & \SI{970}{\gram} \\
Rotor positions & $[\pm 0.142, \pm0.169, 0]$ \\
Motor delay & \SI{0.04}{\second} \\
Rotor thrust curve & $-0.137 + 4.247 \omega + 3.766 \omega^2$  ($\omega \in [0, 1]$)\\
Torque constant & $0.2\frac{\mathrm{N\,m}}{\mathrm{A}}$ \\
Inertia & $\text{diag}([0.008154, 0.005226e, 0.0012043])$ \\ 
\hline
\end{tabular}

\label{table:dynamics-parameters}
\end{table}

Based on this system model, we apply the method for training neural network-based end-to-end policies proposed in \cite{learning-to-fly-in-seconds}. The policy takes the state estimate (position, orientation, linear/angular velocity) and a history of previous actions as the input and maps it to motor-effort setpoints that are directly send to the ESCs (Electronic Speed Controllers). The full training setup can be seen in Fig. \ref{fig:overview} and the policy is directly deployed onto the PX4-based flight controller without an external companion board.

%%%\\\\\\\\\\\\\%%%
%%%\\\\\\\\\\\\\%%%

%%%\\\\\\\\\\\\\%%%
%%%\\\\\\\\\\\\\%%%
%\vspace{3mm}
%\FloatBarrier
%%%\\\\\\\\\\\\\%%%
%%%\\\\\\\\\\\\\%%%
%\vspace{3mm}
%%%%%%%%%%%%%%%%%%%%%%%%%%%%%%%%%%%%%%%%%%%
%%%%%%%%%%%%%%%%%%%%%%%%%%%%%%%%%%%%%%%%%%%
%%%%%%%%%%%%%%%%%%%%%%%%%%%%%%%%%%%%%%%
\section{Experimental Validation} \label{sec:Experimental}
%%%%%%%%%%%%%%%%%%%%%%%%%%%%%%%%%%%%%%%

Experimental validation of the proposed \textit{HoLoArm} drone was conducted across five scenarios.

\begin{itemize}
    \item \textbf{Q1:} Can our drone track trajectories using a RL-based controller?
    
    \item \textbf{Q2:} What is our drone's maximum payload capacity while tracking lemniscate trajectories?

    \item \textbf{Q3:} Is our drone more resilient to in-flight collisions when compared to a rigid drone?

    \item \textbf{Q4:} Can our drone fly through narrow gaps smaller than the width of the aircraft?

    \item \textbf{Q5:} Can our drone survive high velocity impacts during drop tests?

\end{itemize}

%%%%%%%%%%%%%%%%%%%%%%%%%%%%%%%%%%%%%%%%%%%
%%%%%%%%%%%%%%%%%%%%%%%%%%%%%%%%%%%%%%%%%%%

\subsection{RL\_Policy Performance }
%\\\\\\\\\\\\\\\\\\\\\\\\\\\\\\\\\\\\\
%\\\\\\\\\\\\\\\\\\\\\\\\\\\\\\\\\\\\\
%\\\\\\\\\\\\\\\\\\\\\\\\\\\\\\\\\\\\\
\begin{figure}[t]
\centering
\includegraphics[width=0.9\columnwidth]{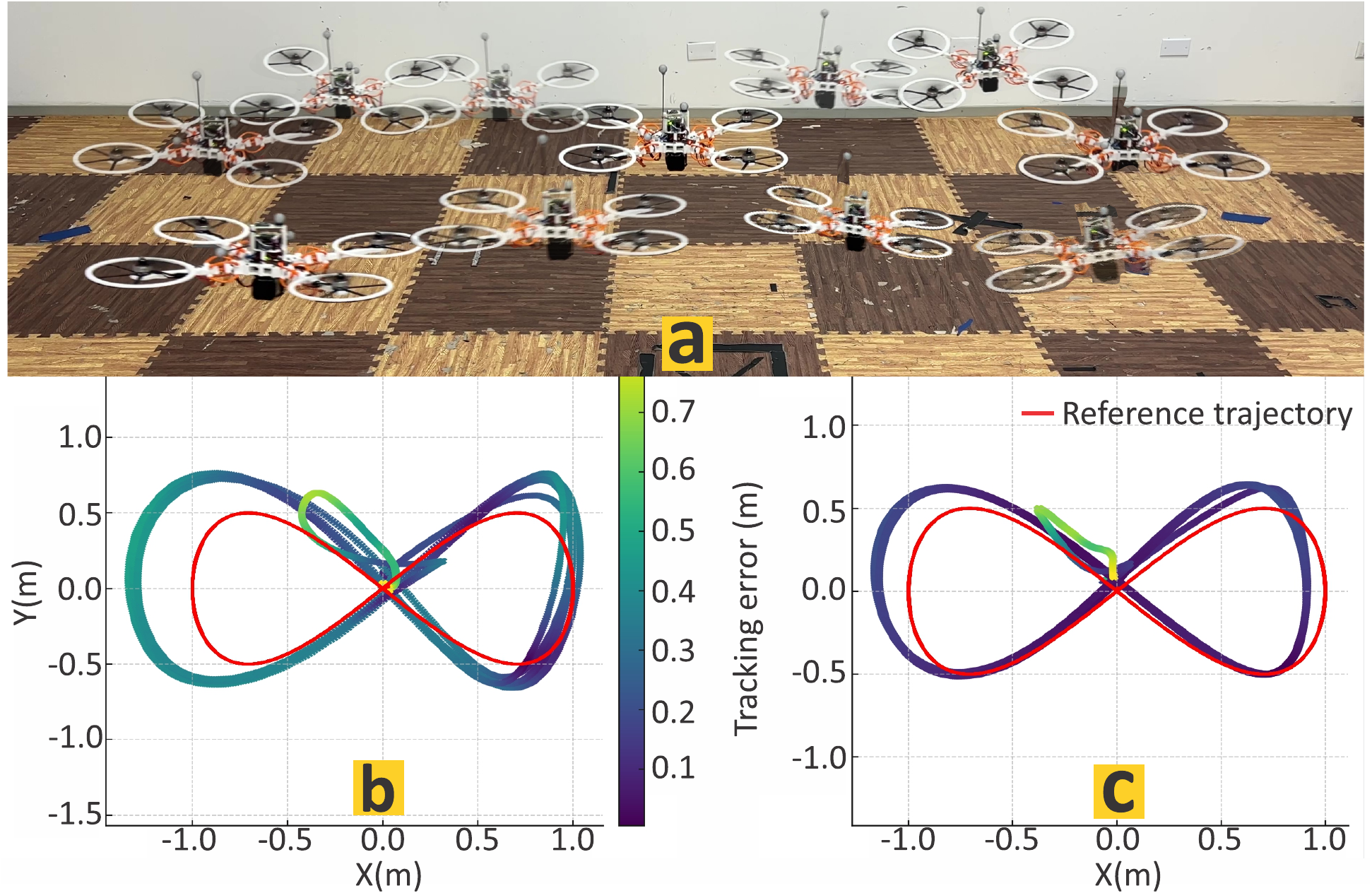}
\caption{Lemniscate testing performance of the controller program: (a) Real-flight image of the drone following a lemniscate trajectory; (b) Slow speed case ($T = 10\,\mathrm{s}$); (c) Normal speed case ($T = 5.5\,\mathrm{s}$).  }
\label{fig:Eight-fingure}
\end{figure}
%\vspace{-2mm}
%\\\\\\\\\\\\\\\\\\\\\\\\\\\\\\\\\\\\
%\\\\\\\\\\\\\\\\\\\\\\\\\\\\\\\\\\\\
%\\\\\\\\\\\\\\\\\\\\\\\\\\\\\\\\\\\\
With the long-term goal of developing a quadrotor capable of actively adapting to collisions in real-world environments, this experiment investigates the trajectory tracking performance of a soft-arm quadcopter design. The proposed flight model achieved mean tracking errors of \(0.52\,\mathrm{m}\) (Normal speed) and \(0.23\,\mathrm{m}\) (Slow speed) are shown in Fig.~\ref{fig:Eight-fingure}. Experiments were conducted in a \(10 \times 6 \times 4\,\mathrm{m}^3\) flying space equipped with a Vicon motion-capture system. For outdoor operations, these error levels are within safe margins for low- to moderate-speed missions without regard to obstacles. Prior work shows that maintaining sub-\(0.5\,\mathrm{m}\) trajectory errors can ensure safe flight when combined with reachability-based safety margins and disturbance-aware control (e.g., control barrier functions under wind) \cite{Kousik2019IROS,Zheng2020CBF,Jiang2024Sensors}. Therefore, our observed mean errors support the feasibility of safe outdoor flight in such scenarios.

%In \ref{fig:Eight-fingure} we compare the tracking performance of the dedicated policy with the foundation policy and find that the dedicated policy performs more smoothly in terms of oscillations (less oscillations in the soft joints) and tracking performance [calculate MSE of each trajectory and put here].

%%%%%%%%%%%%%%%%%%%%%%%%%%%%%%%%%%%%%%%%%%%
%%%%%%%%%%%%%%%%%%%%%%%%%%%%%%%%%%%%%%%%%%%
\subsection {Payload Handling under Complex Trajectories}
%\\\\\\\\\\\\\\\\\\\\\\\\\\\\\\\\\\\\\
%\\\\\\\\\\\\\\\\\\\\\\\\\\\\\\\\\\\\\
%\\\\\\\\\\\\\\\\\\\\\\\\\\\\\\\\\\\\\

\begin{figure}[htbp]
\centering
\includegraphics[width=0.8\columnwidth]{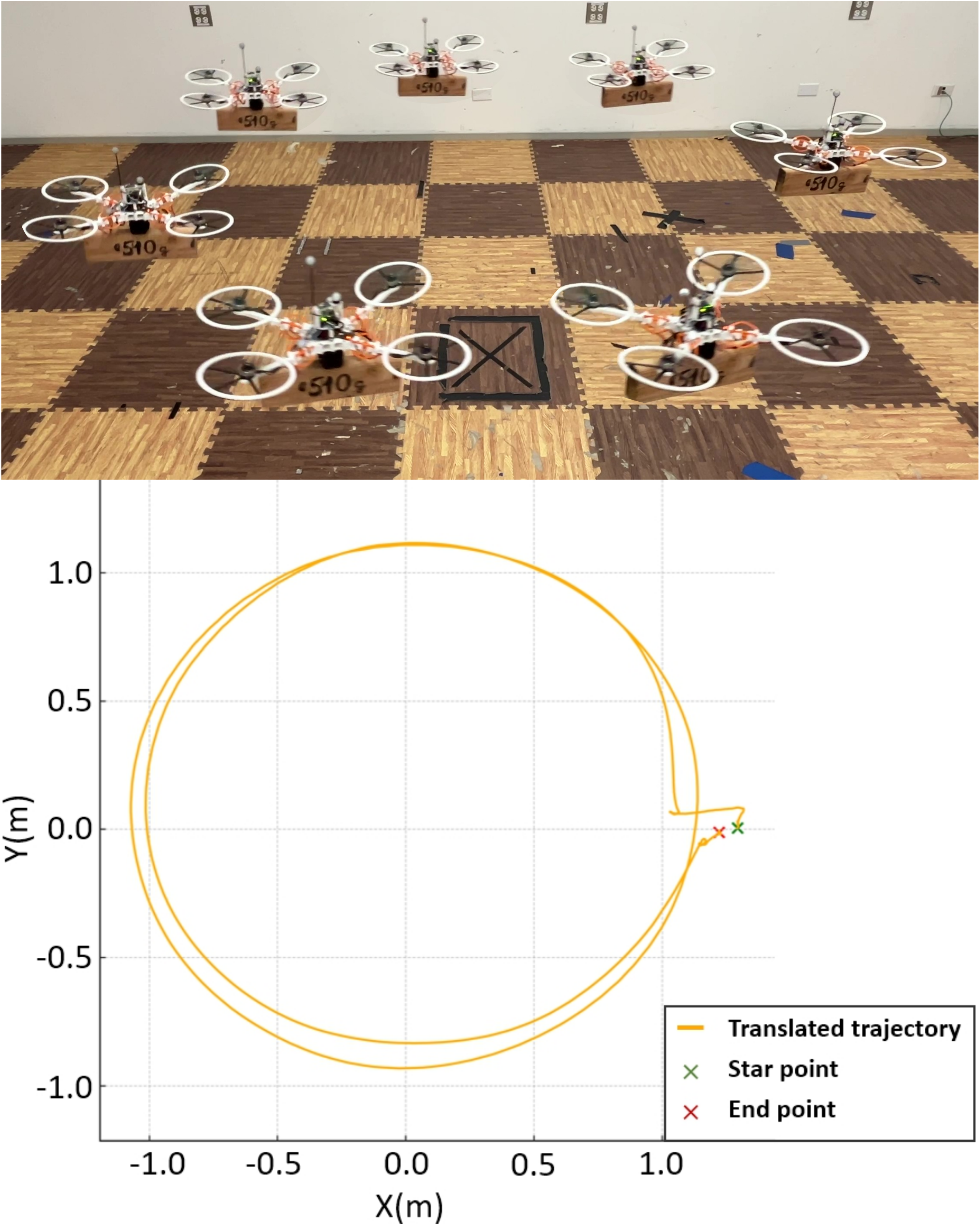}
\caption{Flight experiment showing the drone tracking a circular trajectory at \(90 \mathrm{cm}\) altitude. Trajectory plot showing stable flight of \emph{ HoLoArm} under payload(550\,g), recorded by a 200\,Hz VICON system during a waypoint-tracking task.}
\label{fig: Payload-test-3}
\end{figure}
%\vspace{-2mm}
%\\\\\\\\\\\\\\\\\\\\\\\\\\\\\\\\\\\\
%\\\\\\\\\\\\\\\\\\\\\\\\\\\\\\\\\\\\
%\\\\\\\\\\\\\\\\\\\\\\\\\\\\\\\\\\\\
To demonstrate the practical applicability of the soft-structured \emph{HoLoArm}, we conducted a payload-carrying test where the drone was tasked with making circles while carrying a $550$\,g payload, as illustrated in Fig.~\ref{fig: Payload-test-3}. The drone completed the mission without loss of stability or control. This result highlights the capability of our design to handle moderate payloads, indicating its potential for applications such as lightweight delivery tasks or food transportation, where safe and stable flight is crucial.
%%%%%%%%%%%%%%%%%%%%%%%%%%%%%%%%%%%%%%%%%%%
%%%%%%%%%%%%%%%%%%%%%%%%%%%%%%%%%%%%%%%%%%%
\subsection{Mid-Flight Collision Robustness}
%\\\\\\\\\\\\\\\\\\\\\\\\\\\\\\\\\\\\\
%\\\\\\\\\\\\\\\\\\\\\\\\\\\\\\\\\\\\\
%\\\\\\\\\\\\\\\\\\\\\\\\\\\\\\\\\\\\\

\begin{figure}[htbp]
\centering
\includegraphics[width=0.8\columnwidth]{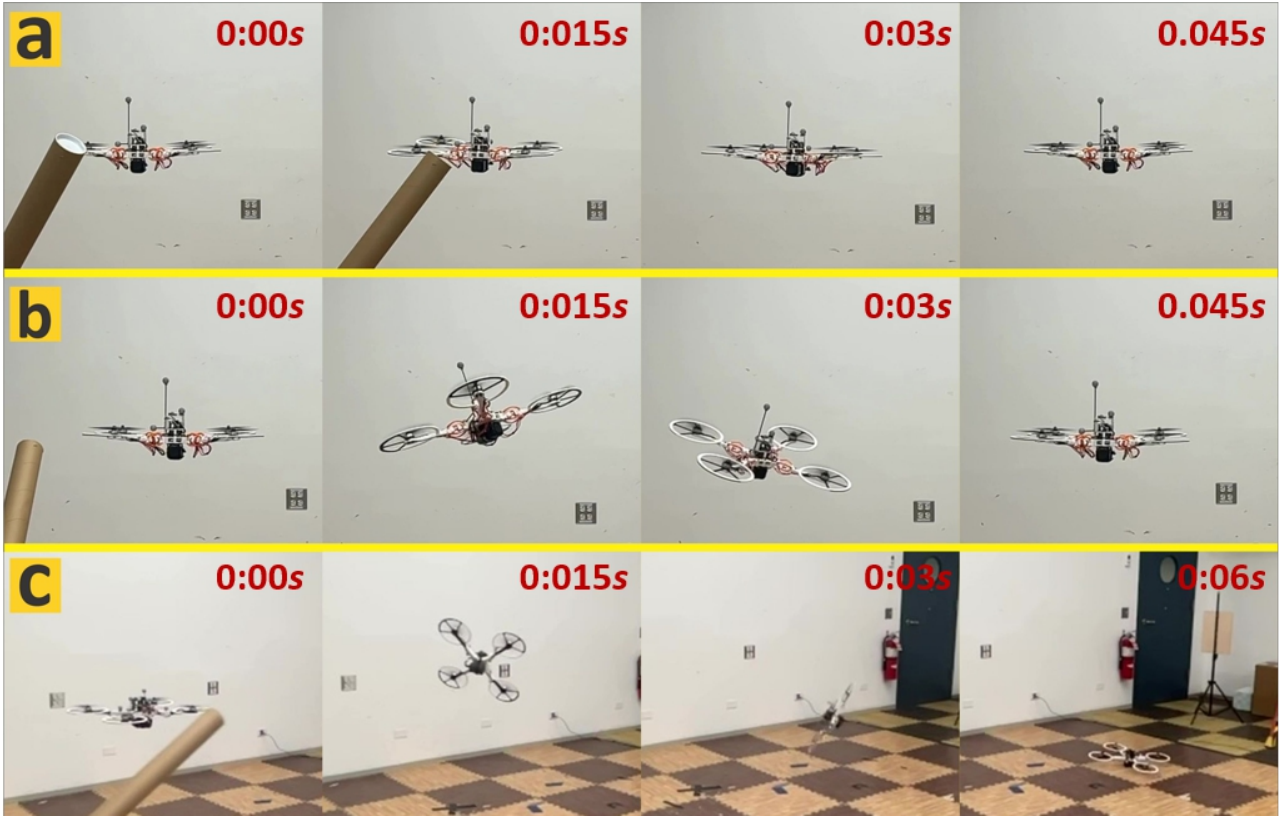}
\caption{Mid-flight collision test: (a,b) The \emph{HoLoArm} drone hovers and stabilize under impact; (c) The rigid body drone drops upon impact.}
\label{fig: Hovering-test}
\end{figure}
%\vspace{-2mm}
In this experiment, we utilized the Vicon motion capture system in combination with the trained \emph{RL\_Policy} to maintain the drone in a stable hovering state. Physical disturbances were then applied to the arm of each drone to evaluate its dynamic response. As illustrated in Fig.~\ref{fig: Hovering-test}(c), the experimental setup is depicted. When subjected to disturbances that induced roll and pitch rotations, the Rigid drone exhibited stronger oscillations compared to the \emph{HoLoArm}, but was still able to regain hovering after each perturbation. However, under disturbances that induced yaw (a direction commonly encountered in real-world flights), the Rigid drone failed to recover and eventually crashed. In contrast, the soft-structured \emph{HoLoArm} drone effectively absorbed yaw-directed forces. Although its yaw angle deviated temporarily in response to the impact, it quickly re-stabilized and maintained its hovering position in mid-air. Figures~\ref{fig: Hovering-test}(a) and (b) show that we attempted to disturb the yaw orientation of the \emph{HoLoArm} drone from both directions. Regardless that, it successfully maintained stability, enabling it to proceed with subsequent challenge tasks without loss of hovering control.

%%%%%%%%%%%%%%%%%%%%%%%%%%%%%%%%%%%%%%%%%%%
%%%%%%%%%%%%%%%%%%%%%%%%%%%%%%%%%%%%%%%%%%%

\subsection{Narrow-Gap Navigation Performance}
%\\\\\\\\\\\\\\\\\\\\\\\\\\\\\\\\\\\\\
%\\\\\\\\\\\\\\\\\\\\\\\\\\\\\\\\\\\\\
%\\\\\\\\\\\\\\\\\\\\\\\\\\\\\\\\\\\\\

\begin{figure}[htbp]
\centering
\includegraphics[width=0.8\columnwidth]{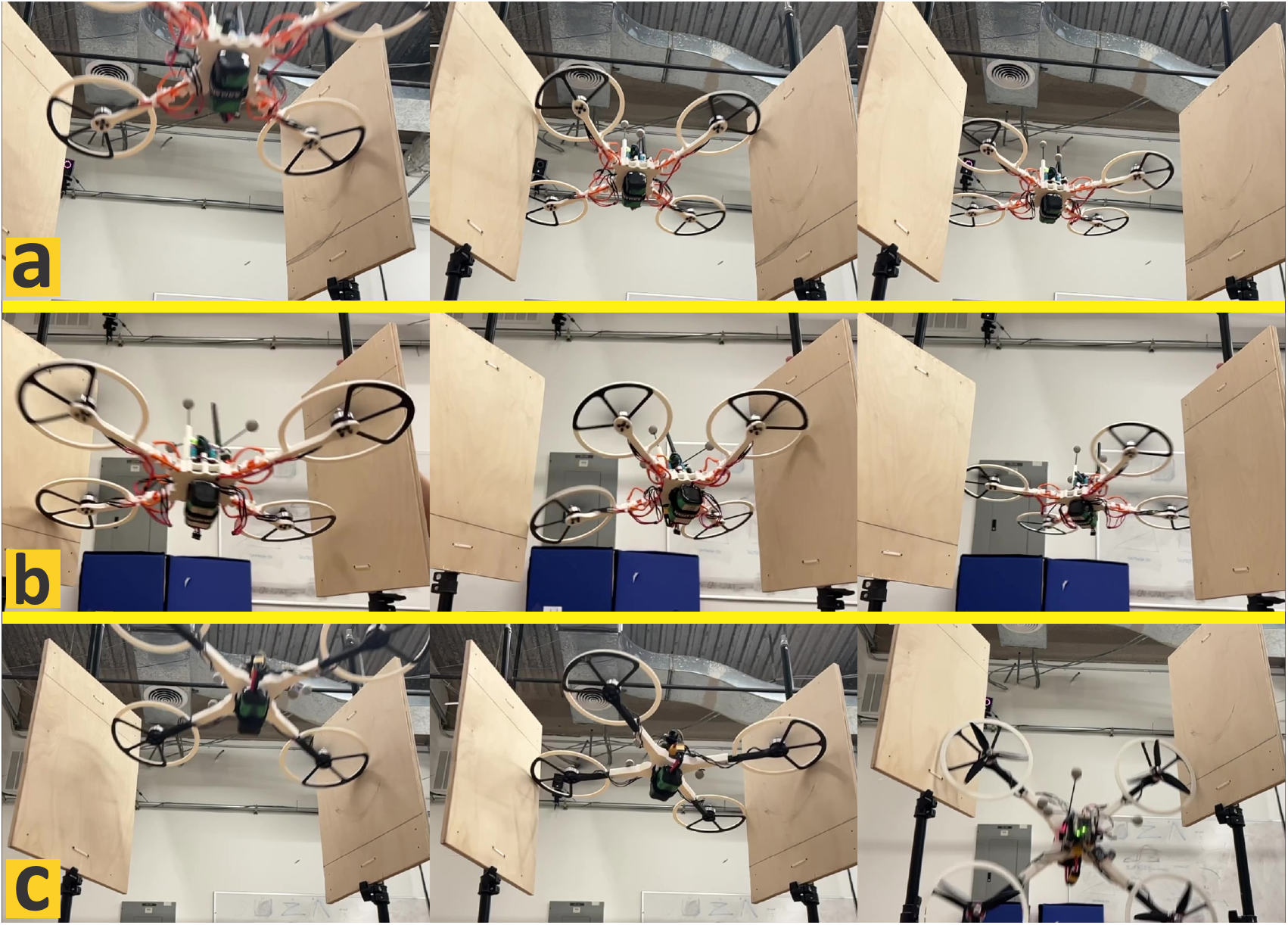}
\caption{Narrow test: (a,b) \emph{HoLoArm} successful with narrow test; (c) A rigid body cannot fly through the same test.}
\label{fig: Narrow-test}
\end{figure}
%\vspace{-2mm}
%\vspace{-5mm}
%\\\\\\\\\\\\\\\\\\\\\\\\\\\\\\\\\\\\
%\\\\\\\\\\\\\\\\\\\\\\\\\\\\\\\\\\\\
%\\\\\\\\\\\\\\\\\\\\\\\\\\\\\\\\\\\\

To demonstrate the morphing capability of the \emph{HoLoArm} drone, we set up a narrow gap whose width was smaller than the nominal lateral dimension of the drone, similar to that in~\cite{DePetris2024Morphy}. In this experiment, both the flexible and rigid drones were tested using an identically configured narrow gap to ensure fairness and transparency in the evaluation. As shown in Fig.~\ref{fig: Narrow-test}(a,b), the \emph{HoLoArm} drone began to deform immediately upon contact between its guards and the designated obstacle surface. After successfully passing through the narrow gap, the drone quickly recovered its original shape and stabilized its flight trajectory. In contrast, the rigid drone was unable to complete the task and was even trapped within the gap, as shown in Fig.~\ref{fig: Narrow-test}(c).
%%%\\\\\\\\\\\\\%%%
%%%\\\\\\\\\\\\\%%%
%\vspace{3mm}
%%%%%%%%%%%%%%%%%%%%%%%%%%%%%%%%%%%%%%%%%%%
%%%%%%%%%%%%%%%%%%%%%%%%%%%%%%%%%%%%%%%%%%%
\subsection{Impact Absorption via Drop Testing}
%\\\\\\\\\\\\\\\\\\\\\\\\\\\\\\\\\\\\\
%\\\\\\\\\\\\\\\\\\\\\\\\\\\\\\\\\\\\\
%\\\\\\\\\\\\\\\\\\\\\\\\\\\\\\\\\\\\\

%\\\\\\\\\\\\\\\\\\\\\\\\\\\\\\\\\\\\
%\\\\\\\\\\\\\\\\\\\\\\\\\\\\\\\\\\\\
%\\\\\\\\\\\\\\\\\\\\\\\\\\\\\\\\\\\\

\begin{figure}[t]
\centering
\includegraphics[width=0.9\columnwidth]{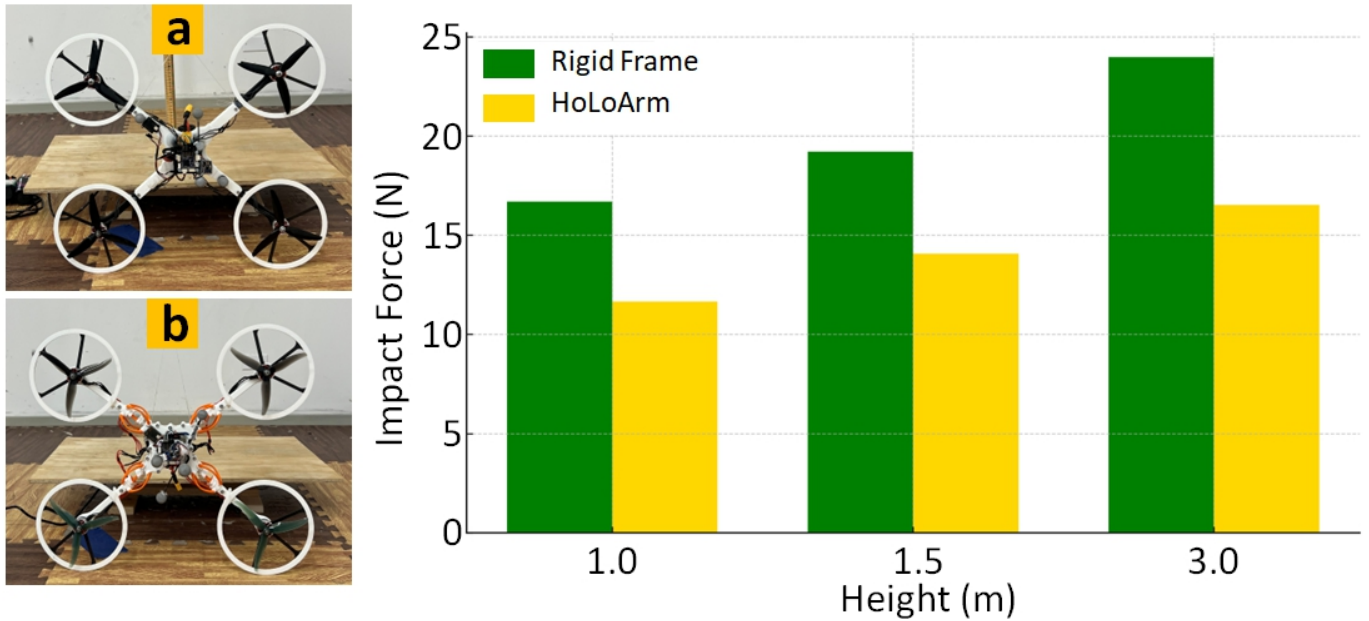}
\caption{Impact force measured at different drop heights during free fall: (a) The rigid drone; (b) The \textit{HoLoArm }.}
\label{fig: Graphic-5}
\end{figure}

In this experiment, we place an electronic scale beneath the impact surface to record the impact force generated by both the Rigid Frame and the \textit{HoLoArm} at various free-fall heights.
Both drones have the same mass of 670\,g and were equally dropped from heights of $1$, $1.5$ and $3$\,m, measured from the drone’s center to the impact surface. Under these initial conditions, the impact velocity ranged from approximately $4.4$\,m/s to $7.7$\,m/s.
At each test height, two free-fall trials were conducted, and the average of the peak force values recorded by the sensor was taken. At the 3 m height, the \emph{Rigid Frame} broke during the first drop and thus could not proceed with the second trial, whereas the \emph{HoLoArm} successfully completed both experimental runs. Fig.~\ref{fig: Graphic-5} illustrates the variation in impact force at each drop height.
The Rigid Frame recorded impact forces of $16.69$\,N, $19.21$\,N, and $23.99$\,N, whereas the \textit{HoLoArm} drone generated significantly lower forces of $11.65$\,N, $14.07$\,N, and $16.52$\,N, respectively. These results highlight the safety performance of the \textit{HoLoArm} when subjected to impact under critical conditions.
%\\\\\\\\\\\\\\\\\\\\\\\\\\\\\\\\\\\\\
%\\\\\\\\\\\\\\\\\\\\\\\\\\\\\\\\\\\\\
%\\\\\\\\\\\\\\\\\\\\\\\\\\\\\\\\\\\\\

%\vspace{-2mm}

\begin{figure}[t]
\centering
\includegraphics[width=0.8\columnwidth]{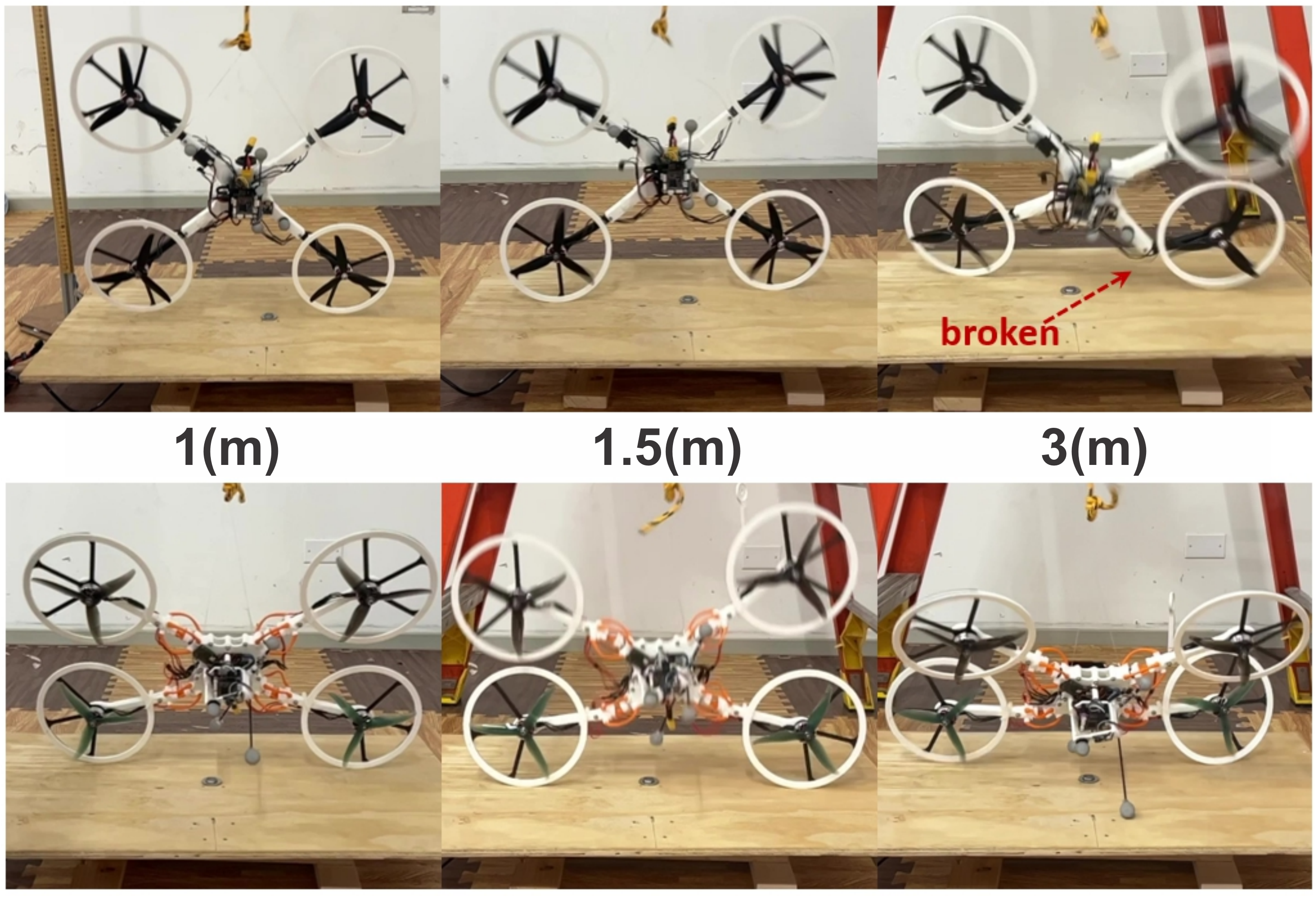}
\caption{Shape variation during free fall from different heights. 
The rigid drone (top) broke upon impact from a height of 3\,m, 
while the \textit{HoLoArm} (bottom) remained completely intact.}
\label{fig: Drop-1m3m}
\end{figure}
%\vspace{-2mm}
%\\\\\\\\\\\\\\\\\\\\\\\\\\\\\\\\\\\\
%\\\\\\\\\\\\\\\\\\\\\\\\\\\\\\\\\\\\
%\\\\\\\\\\\\\\\\\\\\\\\\\\\\\\\\\\\\
In addition to force data, Fig.~\ref{fig: Drop-1m3m} visualizes the morphological changes observed in the two drones after impact.
The \textit{HoLoArm} exhibits increasing deformation with higher drop heights.
In contrast, the Rigid Frame shows minimal visible change at 1.5\,m, but suffers catastrophic failure at 3\,m and is unable to recover.
Meanwhile, the \textit{HoLoArm } remains fully intact after the same drop, demonstrating its enhanced resilience.

% It turns out that a stiff robot link did not benefit much from a kinematic reactive controller due to its well-known low bandwidth characteristic.

%%%%%%%%%%%%%%%%%%%%%%%%%%%%%%%%%%%%%%%
\section{Conclusion and Future Works} \label{Conclution}
%%%%%%%%%%%%%%%%%%%%%%%%%%%%%%%%%%%%%%%

This work presented the design and fabrication of a bio-inspired drone equipped with compliant arms. Experimental results demonstrated that the proposed \textit{HoLoArm }effectively reduces multi-directional impacts during flight. The integration of compliant structures at critical locations enhances safety for both the drone and its surroundings, with impact forces significantly lower than those of conventional rigid drones. Furthermore, the Reinforcement Learning (RL) policy proved effective, enabling the controller to quickly adapt to the more complex hardware configuration of the \textit{HoLoArm }, allowing it to achieve pre-planned flight tasks without relying on a detailed dynamic model. These findings highlight the potential for developing high-performance aerial platforms with superior collision resilience in human-centered environments.

Future works will focus on three directions. First, we aim to develop a high-fidelity mathematical model of the compliant structure using analytical methods, enabling model-based optimal control design. Second, we intend to realize a fully compliant drone architecture, including the previously published soft propellers, integrated with safety-oriented control algorithms for safe operation in shared spaces. Third, we will investigate material properties and structural configurations to establish a \textit{scaling law}, allowing the proposed design to be adapted to various frame sizes depending on the intended application.

%%%%%%%%%%%%%%%%%%%%%%%%%%%%%%%%%%%%%%%
%\section*{Acknowledgment}
%%%%%%%%%%%%%%%%%%%%%%%%%%%%%%%%%%%%%%%

\bibliographystyle{IEEEtran}
\bibliography{references}

\end{document}